\documentclass[letter, journal]{IEEEtran}
\ifCLASSOPTIONcompsoc
  \usepackage[nocompress]{cite}
\else
  \usepackage{cite}
\fi
\ifCLASSINFOpdf
\else
\fi
\usepackage{hyperref}
\usepackage{algorithm}
\usepackage{algpseudocode}
\usepackage{graphicx}
\usepackage{booktabs}
\usepackage{colortbl}
\usepackage{tikz}
\usetikzlibrary{patterns}
\usepackage{pgfplots}
\usepackage{pgfplotstable}
\usepackage{newfloat}
\usepackage{listings}
\usepackage{amsmath,amssymb,amsfonts}
\usetikzlibrary{arrows.meta,
                chains,
                positioning}
\usepackage{subcaption}
\usepackage{booktabs}
\usepackage{multirow}

\begin{document}
\title{PDRL: Multi-Agent based Reinforcement Learning for Predictive Monitoring}

\author{Thanveer Shaik,  Xiaohui Tao,  Lin Li, Haoran Xie, U R Acharya, Raj Gururajan, Xujuan Zhou
% \thanks{This paragraph of the first footnote will contain the date on 
% which you submitted your paper for review. It will also contain support 
% information, including sponsor and financial support acknowledgment. For 
% example, ``This work was supported in part by the U.S. Department of 
% Commerce under Grant BS123456.'' }
\thanks{Thanveer Shaik, Xiaohui Tao are with 
the School of Mathematics, Physics \& Computing, University of Southern Queensland, Toowoomba, Queensland, Australia (e-mail: Thanveer.Shaik@usq.edu.au, Xiaohui.Tao@usq.edu.au).}
\thanks{Lin Li is with the School of Computer and Artificial Intelligence, Wuhan University of Technology, China (e-mail: cathylilin@whut.edu.cn)}
\thanks{Haoran Xie is with the Department of Computing and Decision Sciences, Lingnan University, Tuen Mun, Hong Kong (e-mail: hrxie@ln.edu.hk)}

\thanks{U R Acharya is with School of Mathematics, Physics \& Computing, University of Southern Queensland, Toowoomba, Queensland, Australia (e-mail: Rajendra.Acharya@usq.edu.au).}
\thanks{Raj Gururajan is with School of Business, University of Southern Queensland, Springfield, Queensland, Australia  (e-mail: Raj.Gururajan@usq.edu.au).}
\thanks{Xujuan Zhou  is with School of Business, University of Southern Queensland, Springfield, Queensland, Australia  (e-mail: Xujuan.Zhou@usq.edu.au).}
}

\IEEEtitleabstractindextext{%
\begin{abstract}
Reinforcement learning has been increasingly applied in monitoring applications because of its ability to learn from previous experiences and can make adaptive decisions. However, existing machine learning-based health monitoring applications are mostly supervised learning algorithms, trained on labels and they cannot make adaptive decisions in an uncertain complex environment. This study proposes a novel and generic system, predictive deep reinforcement learning (PDRL) with multiple RL agents in a time series forecasting environment. The proposed generic framework accommodates virtual Deep Q Network (DQN) agents to monitor predicted future states of a complex environment with a well-defined reward policy so that the agent learns existing knowledge while maximizing their rewards. In the evaluation process of the proposed framework, three DRL agents were deployed to monitor a subject's future heart rate, respiration, and temperature predicted using a BiLSTM model. With each iteration, the three agents were able to learn the associated patterns and their cumulative rewards gradually increased. It outperformed the baseline models for all three monitoring agents. The proposed PDRL framework is able to achieve state-of-the-art performance in the time series forecasting process. The proposed DRL agents and deep learning model in the PDRL framework are customized to implement the transfer learning in other forecasting applications like traffic and weather and monitor their states. The PDRL framework is able to learn the future states of the traffic and weather forecasting and the cumulative rewards are gradually increasing over each episode.
\end{abstract}

}

\maketitle

% is not selected <OR> if conference mode is selected - because compsoc
% conference papers position the abstract like regular (non-compsoc)
\IEEEdisplaynontitleabstractindextext

% \fi
\IEEEpeerreviewmaketitle

\ifCLASSOPTIONcompsoc
\IEEEraisesectionheading{\section{Introduction}\label{sec:introduction}}
\else
\section{Introduction}
\label{sec:introduction}
\fi

Data mining has been widely adopted for analysis and knowledge discovery in databases. This process involves data management, data preprocessing, modeling, and results in inferences and extracting latent data patterns~\cite{Shu2022}. Early warning systems based on data mining have enabled applications to perform a risk analysis, monitoring and warning, and a response capability~\cite{Talari2022}. Using existing knowledge or a set of indicators can assist domains such as healthcare~\cite{Shaik2022} and education~\cite{Hu2022} to design decision support systems. Radanliev et al.~\cite{Radanliev2020} study used data mining to investigate scientific research response to the COVID-19 pandemic and to review key findings on how early warning systems developed in previous epidemics responded to contain the virus. 

Traditional unsupervised learning techniques discover underlying patterns for knowledge discovery in unlabelled data using association rule mining and clustering techniques~\cite{Pattanayak2022}. Supervised learning strategies learn from labeled data to classify or predict patients' physical activities and vital signs~\cite{Thirunavukarasu2022}. However, these methodologies are highly dependent on data and can only observe the data and present possible decisions in response, they cannot take actions based on observations. Reinforcement learning (RL) deploys a learning agent in an uncertain, complex environment that explores or exploits the environment with its actions and learns the data based on its experience~\cite{9868114,9950249}. This allows the learning agent to gain rewards based on learning and its actions.

\begin{figure}[!ht]
    \centering
    \includegraphics[width=\columnwidth]{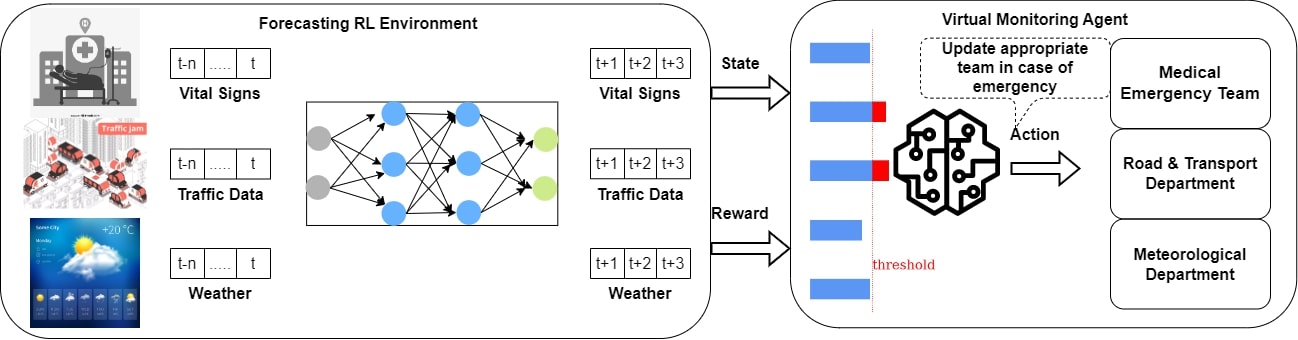}
    \caption{PDRL framework to monitor different forecasting applications and alert appropriate emergency teams.}
    \label{fig:graph_abs}
\end{figure}

% With its ability to deal with sampled, delayed, and evaluative feedback at the same time, RL can be adopted for diagnosing decisions or dynamic treatment regimes in which feedback is delayed.

RL is used in dynamic domains such as stockmarket trading~\cite{ZHANG2022108543,ZHANG2022108490}, traffic prediction~\cite{han2021dynamic,fang2021spatial}, and weather forecast~\cite{kumar2021micro} to tackle decision-making problems using agent-environment interaction samples and potentially delayed feedback~\cite{10.1145/3477600} that could also be applied to healthcare applications. In the healthcare domain, chronically diseased such as Parkinson's disease~\cite{8520887} and critical care patients often require long-term dynamic treatment regimes with the timely intervention of clinicians to avoid unwanted outcomes~\cite{hong2022state}. Zeng et al.~\cite{zeng2022optimizing} proposed an RL algorithm to optimize post-operation warfarin anticoagulation dosage. The RL results outperformed  conventional clinical practice using rule-based algorithms. Existing patient monitoring applications based on RL primarily focus on prescribing the timing and dosage of medications~\cite{Watts2020} so that patients are administered take the right medication at the right time~\cite{Naeem2021}. Chen et al.~\cite{Chen2021} described Probabilistic machine learning models such as RL using the analogy of an ICU clinician (learning agent) to monitor a patient (environment) via actions like ventilation and observing the changes in the environment (patient's state) to make subsequent decisions that achieve the goal of discharging the patient successfully.

The research problem addressed here is that of being able to monitor the predicted state of an environment and take appropriate actions to avoid an emergency. Traditional supervised learning strategies can classify or predict based on their training but cannot monitor and alert the appropriate team for timely interventions. To assist with tracking the environment state and monitor certain parameters we have designed a virtual generic forecasting environment with observation space, actions, and rewards policy with multiple deep learning agents. Deploying a single learning agent to monitor all the parameters would complicate the environment as there are different thresholds set up for each of the parameters in an environment. For example, the learning agents learn different threshold levels of each vital sign in modified early warning scores (MEWS)~\cite{signscanberra} based on previous iterations and rewards being accumulated for its actions. Well-trained RL agents are capable of monitoring a patient's vital signs such as heart rate, respiration rate, and temperature, and alerting the corresponding clinical team if the vital signs fall outside any of the predefined thresholds~\cite{Chen2022}.

Modeling forecasting applications, such as vital signs prediction, traffic prediction, and weather prediction, as an RL environment can enable RL agents to monitor tasks. RL agents can learn from historical data and interact with the environment to make real-time decisions based on the predicted states or actions. This approach can be used to develop intelligent monitoring systems that can adapt to changing conditions, optimize actions, and make informed decisions in complex and dynamic environments. By using RL for monitoring, it is possible to automate and optimize monitoring processes in various domains, leading to more efficient and effective monitoring outcomes. In this study, the RL environment is configured with a deep learning model to predict future states, which are then monitored by an RL agent.

The aim of this research is to create a multi-agent framework that utilizes deep reinforcement learning (DRL) agents to monitor and learn data patterns for various parameters. Each parameter will have its own DRL agent, responsible for monitoring, learning, and alerting respective teams if the parameters deviate from predefined thresholds as shown in Fig.~\ref{fig:graph_abs}. Conventional RL methodology is an agent performing a task for a transition from one state to another state, where this action might reward the agent either positively or negatively. In this study, a novel approach is taken to assign rewards so that the RL agents learn data patterns. An agent gets rewarded for predicting an action and performing the action in its current state. The rewards are designed in such a way that the learning agents are penalized for predicting the wrong actions. To learn behaviors we follow the Reward-is-enough hypothesis~\cite{silver2021reward} being that the learning agent always tries to maximize the rewards based on their previous actions.  The contributions of this study are as follows:
\begin{itemize}
    \item A generic monitoring environment accommodates multiple agents to monitor the states of a  forecasting environment.
    \item Proposed a model-free gaming agent to learn the existing knowledge and monitor underlying data patterns adaptively.
    \item Transfer learning approach for time series forecasting applications such as patients' health status, traffic, and weather using the multi-agents in the PDRL environment.
\end{itemize}

The paper is organized as follows: Section~\ref{relatedworks} presents the related works in the RL community to learning human behavior patterns and application in the healthcare domain. Research  problem formulation and the proposed multi-agent PDRL framework have been detailed in Section~\ref{methods}. In Section~\ref{experiment}, the proposed methodology is evaluated on 10 different subject vital signs, and baseline models are discussed. In Section~\ref{results}, the results of the proposed approach are compared with baseline models, and hyper-parameter optimization of the learning rate and discount factor are discussed. The comparison between the supervised approach and the RL approach is discussed in Section~\ref{discussion}. Section~\ref{conclusion} concludes the paper with limitations and future work.

\section{Related Works}\label{relatedworks}
\subsection{Data Mining in Early Warning Systems}
Ak{\c{c}}ap{\i}nar et al.~\cite{Akapnar2019} proposed a study that uses interaction data from online learning to predict the academic performance of students at end of term by using the kNN algorithm which predicted unsuccessful students at an 89\% rate. It also suggests that performance can be predicted in 3 weeks with 74\% accuracy, useful for early warning systems and selecting algorithms for analysis of educational data. Cano et al.~\cite{Cano2019} presented a multiview early warning system for higher education that uses comprehensible Genetic Programming classification rules to predict student performance, specifically targeting underrepresented and underperforming student populations. The authors integrated various student information sources and have interfaces to provide personalized feedback to students, instructors, and staff. In healthcare, Hussain-Alkhateeb et al.~\cite{HussainAlkhateeb2021} conducted a scoping review to summarize existing evidence of early warning systems for outbreak-prone diseases such as chikungunya, dengue, malaria, yellow fever, and Zika. It found that while many studies showed the quality performance of their prediction models, only a few presented statistical prediction validity of early warning systems. It also found that no assessment of the integration of the early warning systems into a routine surveillance system could be found and that almost all early warning systems tools require highly skilled users with advanced statistics. Spatial prediction remains a limitation with no tool currently able to map high transmission areas at small spatial levels. Liu et al.~\cite{Liu2021} conducted a study on the use of data mining technology to analyze college students' psychological problems and mental health. The authors use intuitionistic fuzzy reasoning judgment, analytic hierarchy process, and expert scoring method to construct a model for studying college students' online public opinion and use data mining techniques such as the decision tree algorithm and Apriori algorithm to analyze students' psychological problems and provide decision-making support information for the school psychological counseling center. In \cite{Moghadas2020}, the authors talk about the challenges of managing large amounts of data from connected devices and how data mining can be used to extract valuable information. It also mentions the use of fog computing technology to improve the quality of service in healthcare applications. The article suggests wearable clinical devices for continuous monitoring of individual health conditions as a solution for chronic patients. An EWS (Early Warning System) for heavy precipitation using meteorological data from Automatic Weather Stations (AWSs) is proposed by Moon et al.~\cite{Moon2019} and its performance are measured by various criteria. 
% It is important to have accurate and timely warning information to minimize the damage caused by heavy rainfall. While EWSs for water-related hazards such as floods and landslides have been reviewed and tested, there has been no quantitative performance evaluation of EWSs for heavy rainfall. 
% The paper proposes a machine-learning method to forecast rainfall and predicts the expected amount of precipitation in millimeters.

\subsection{RL Monitoring}
In the healthcare domain, Lisowska et al.\cite{lisowska2021personalized} developed a digital behavior-change intervention to help cancer patients build positive health habits and enhance their lifestyles. The authors used reinforcement learning to learn the appropriate time to send the intervention prompts to the patients. Furthermore, effective prompt policies to perform an activity have been used in a custom patient environment. Three RL approaches Deep Q-Learning (DQL), Advance Actor-Critic (A2C) and proximal policy optimization (PPO) have been used to suggest a virtual coach for sending a prompt. Similarly, personalized messages enhance physical activity in type 2 diabetes patients~\cite{yom2017encouraging}. Li et al.\cite{li2022electronic} proposed an electronic health records (EHRs)-based reinforcement learning approach for sequential decision-making tasks. The authors used a model-free DQN algorithm to learn the patients' data and provide clinical assistance in decision-making. Co-operative multi-agent RL has been deployed using value compositions and achieved better results. RL decision-making ability can be used for human activity recognition as per~\cite{guo2021deep}. The authors proposed a dynamic weight assignment network architecture in which twin delayed deep deterministic (TD3)~\cite{fujimoto2018addressing} algorithm was inspired by Deep Deterministic Policy Gradient algorithm (DDPG), Actor-Critic, and DQN algorithms. RL agents tend to learn effective strategies while the sequential decision-making process using trial-and-error interactions with their environments~\cite{10.1145/3477600}.

\subsection{Mimic Human Behavior Patterns}
Tirumala et al.\cite{tirumala2020behavior} discussed learning human behavior patterns and capturing common movement and interaction patterns based on a set of related tasks and contexts. The authors discussed probabilistic trajectory models to learn behavior priors and proposed a generic framework for hierarchical reinforcement learning (HRL) concepts. Janssen et al.\cite{janssen2022hierarchical} suggested breaking a complex task such as biological behavior into more manageable subtasks. HRL is able to combine sequential actions into a temporary option. The authors discussed how biological behavior is conceptually analogous to options in HRL. Tsiakas et al.\cite{tsiakas2017interactive} proposed a human-centric cyber-physical systems (CPS) framework for personalized human-robot collaboration and training. This framework focuses on monitoring and assessment of human behavior. Based on the multi-modal sensing framework, the authors aimed for effective human attention prediction with a minimal and least intrusive set of sensors. Kubota et al.~\cite{kubota2022methods} examined how robots can adapt the behavior of people with cognitive impairments as part of cognitive neuro-rehabilitation. A variety of robots in therapeutic, companion, and assistive applications has been explored in the study. 
% High-level human behaviors such as cognitive abilities, and engagement with low-level behaviors such as speech, gesture, and physiological signs need to be passed to robots to perceive and understand human behavior~\cite{pmlr-v173-barquero22b}, especially for health applications. 

In addition to RL applications in the gaming industry, a great deal of research is being conducted using RL to learn and mimic human behavior and also to deploy socially assistive robots. However, physical robots with human interaction capability cannot be deployed at sensitive locations like hospitals, educational institutions, elderly home care, and mental health facilities as they might cause safety issues for students, patients, carers, and medical staff~\cite{Almohri2017}. Existing monitoring applications with supervised or unsupervised learning cannot cope with uncertain events in a dynamic and complex environment but the supervised approach is well known for its achievement in regression problems. Virtual robots with adaptive learning abilities can be deployed to overcome these issues to monitor and learn the predicted states from a supervised learning model. In this study, we developed a custom monitoring environment to learn behavior patterns from predicted states by designing rewards according to the applications in certain domains. The framework is capable of alerting the appropriate team based on the level of severity and assisting in timely intervention. 
% This work takes a novel approach to learning behavior patterns and monitoring them.

\begin{figure*}
    \centering
    \includegraphics[width=0.85\textwidth]{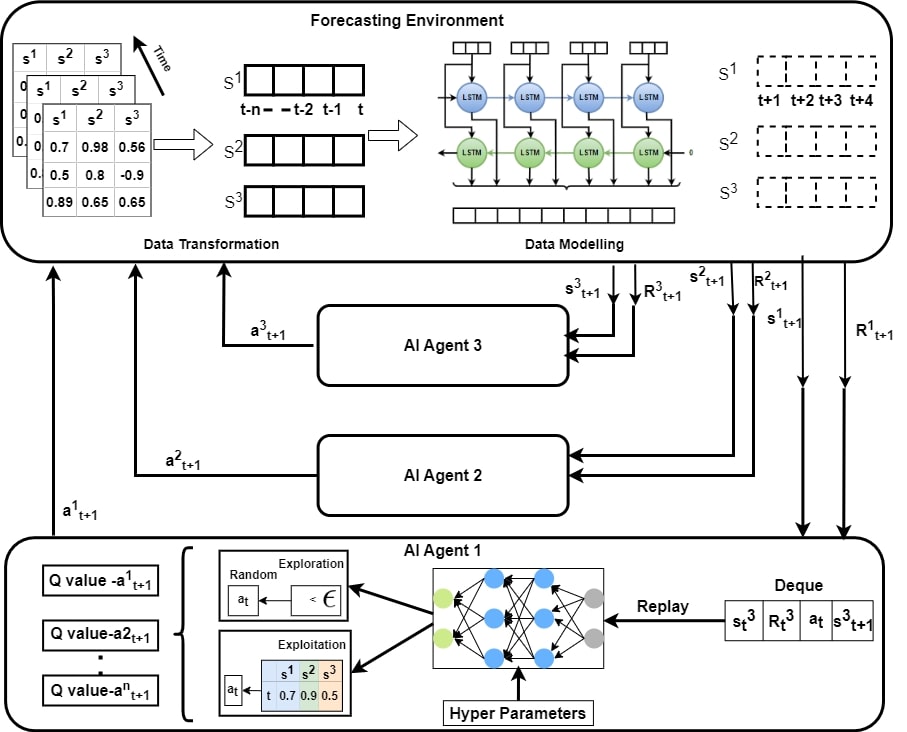}
    \caption{PDRL monitoring Framework}
    \label{fig:method}
\end{figure*}

\section{PDRL Monitoring Framework}\label{methods}
In this section, custom behavior forecasting RL environment and monitoring agents of the proposed predictive deep reinforcement learning(PDRL) framework has been discussed in detail along with problem formulation. This study is to learn data patterns in an uncertain environment by monitoring its current state, taking appropriate actions, and getting rewarded for the actions as shown in Fig.~\ref{fig:method}.  

\subsection{Problem Formulation}
Knowledge discovery in data mining can be achieved by utilizing previously known and potentially useful information extracted from observations and various data mining techniques. The research problem involves designing a multi-agent framework to monitor data in a forecasting environment and uncover underlying patterns in relation to thresholds established by known knowledge. For example, consider a scenario where a client, $c_{n} \epsilon C$ where $n=1,2,3..N$, $C=|N|$ is the number of clients with wearable sensors on their bodies to track and forecast vital signs. Each client $c$ has a set of DRL learning agents to monitor predicted vital signs and alert the appropriate emergency team if health parameters exceed modified early warning scores (MEWS)~\cite{signscanberra}.

To formulate this problem, a customized RL forecasting environment needs to be configured with an innovative reward policy that links the current state and agent actions to learn data patterns while maximizing their rewards. This can be achieved based on a Markov Decision Process (MDP), which can be defined as a 5-tuple $M$= (S,A,P,R,$\gamma$), where: $S$ is a finite state space, with $s_{t}\epsilon S$ denoting the state of an agent at time $t$, $A$ is a set of actions defined for the agent, with $a_{t}\epsilon A$ denoting the action taken by the agent at time $t$, $P$ is a Markovian transition function as shown in Equation~\ref{markov}, which denotes how the agent transits from state $s$ to state $s'$ while performing an action $a$, $R$ is a reward function, which returns an immediate reward $R(s,a)$ for the action $a$ taken in a state $s$ defined in Equation~\ref{reward}, ${\gamma}$ is a discount factor that focuses on immediate rewards instead of future rewards. It remains between 0 and 1.

\begin{equation}\label{markov}
    {\displaystyle P_{a}(s,s')=\Pr(s_{t+1}=s'\mid s_{t}=s,a_{t}=a)}
\end{equation}
\begin{equation}\label{reward}
    {\displaystyle R(s_{t},a_{t})=\sum _{t=0}^{\infty }\gamma ^{t}r_{t},}
\end{equation}

The next step of the research problem is to compute the optimal reinforcement learning policy $\pi: S \times A \rightarrow [0,1]$, which helps in predicting the probability that an agent selects an appropriate action $a_{t}\epsilon A$ in a specific state $s_{t} \epsilon S$ at time $t$. To do this, the action value (Q-function) needs to be updated in each iteration and can be defined in Equation~\ref{optimal-bellman}. $Q^{new}(s_{t},a_{t})$ is the new output of the action $a_{t}$ and state value $s_{t}$. $\alpha$ is the learning rate, which determines how much information from the previously computed Q-value is used for the given state-action pair. $\gamma$ is a discount factor that focuses on immediate rewards instead of future rewards, and it remains between 0 and 1.

\begin{equation}\label{optimal-bellman}
{\displaystyle Q^{new}(s_{t},a_{t})\leftarrow (1-{\alpha}) {Q(s_{t},a_{t})}+ \alpha \cdot {{ (}{{r_{t}}+{\gamma}{\max _{a}Q(s_{t+1},a)}} )}}
\end{equation}

\subsection{Forecasting Environment}\label{Environment} 
In this section, forecasting applications are modeled as a customized RL forecasting environment based on MDP has been designed with observation space $S$, and action space $A$ for learning agents to take appropriate actions, and it rewards $R$ for the agents' actions. The forecasting environment enables a deep learning model to forecast the future states at time $t+1,t+2,t+3,t+4$ based on the training data at previous timesteps $t-n,...,t-2,t-1,t$ in the proposed PDRL framework as shown in Fig.~\ref{fig:method}. 

\subsubsection{\textbf{Forecasting States}}
In this study, forecasting the future states of an environment is a supervised time series learning approach. For this task, the recurrent neural network (RNN) model variant, the bidirectional LSTM(Bi-LSTM) model is deployed. Mathematically, the Bi-LSTM model is defined in Equation~\ref{lstm-eq}. A regularization method, dropout~\cite{hassan2021end} was used to exclude activation and weight updates of recurrent connections from LSTM units probabilistically.

\begin{equation}\label{lstm-eq}
\begin{split}
y(x) = \sum_{i=1}^{n} Activation1(b+w_{i} x_{i})\\
Bi-LSTM(y) = Activation2(\frac{e^{y_{i}}}{\sum _{j=1}^{k} e^{y_{j}}})
\end{split}
\end{equation}

where~$b$: Bias added on each hidden layer,$x$: Input value, $w$: Weights added on each hidden layer, $y$: Output value from each neuron, $Activation1$: Activation functions on input and hidden layers, $Activation2$: Activation function on the output layer.

Based on the forecasted states, the following components are configured in the forecasting RL environment.

\subsubsection{\textbf{Observation Space}}
The environment shown in Fig.~\ref{fig:method} has state $s_{t}^{i} \epsilon S$ where $i=0,1,2,...n$, observations in a state at time t. The idea is to split the state into observations and forecast the states based on the time series data. The predicted states are getting assigned to multi-agents. Furthermore, considering a single agent to monitor the multiple states of a complex environment might lead to a sparse rewards challenge where the environment rarely produces a useful reward and limits agent learning. Hence, multiple agents need to be deployed to monitor multiple states. To determine the expected return $E_{\pi}$ of a policy $\pi$ in a state $s$ can be defined in state-value Equation~\ref{modfied_value_function} adopting multi-agent where $i=0,1,2,3,...n$ is a finite number of observations $n$ in a state.

\begin{equation}\label{modfied_value_function}
V^\pi (s^{i}) = E_{\pi}\biggl\{\ \sum _{t=0,i=0}^{\infty,n} \gamma^{t}R(s_{t},\pi(s_{t}))| s_{0}^{0}=s\biggl\}
\end{equation}

\subsubsection{\textbf{Action Space}}
Defining actions for the RL agent in the environment is the most critical part of the RL process as it directly reflects the capacity of RL agents in adaptive learning. In this study, a discrete set of actions are proposed for a continuous observation space. Each of these actions will be chosen by agents based on the current state of the forecasting environment. The expected return $E_{\pi}$ for taking an action $a$ in a state $s$ under a policy $\pi$ can be measured using action-value function $Q _{\pi}(s,a)$ Equation~\ref{modified_action_value}.

\begin{equation}\label{modified_action_value}
Q^{\pi}(s,a)  = E_{\pi}\biggl\{\ \sum _{t=0}^{\infty} \gamma^{t}R(s_{t},a_{t},\pi(s_{t}))| s_{0}=s, a_{0}=a \biggl\}
\end{equation}

Actions within an RL environment vary depending on the application. For instance, in a health monitoring application, the patient's health status may change based on vital signs such as heart rate, blood pressure, respiratory rate, and more. Actions can be configured based on these vital signs, utilizing modified early warning scores~\cite{signscanberra}. Threshold levels defined for vital signs, such as heart rate, can be used to measure the level of emergency, and appropriate alerts can be triggered accordingly. For example, if the heart rate exceeds a predefined threshold, a high-level emergency alert may be generated, while a lower-level alert may be issued for a heart rate that is moderately deviating from the normal range. This way, RL agents can take actions based on the defined thresholds to ensure timely and appropriate responses in health monitoring applications.

\subsubsection{\textbf{Rewards}}
RL goals can be represented by cumulative rewards achieved by learning agents with their actions in an environment. Conventional reinforcement learning rewards an agent based on their action for a transition from a state $s_{t}$ to $s_{t+1}$. In this study, the goal of the learning agent is to learn underlying data patterns in terms of states of a forecasting environment and this can be achieved by the efficient design of a reward policy. The reward policy defined for this study is calculated using Equation~\ref{reward_cal} where agents get positively rewarded only if an agent monitors the state and predicts the right action from the action space. Otherwise, the agent will be negatively rewarded. 

\begin{equation}\label{reward_cal}
{R(s_{t},a_{t}) =\begin{cases}
+reward &\text{if $action$ is appropriate}\\
-reward &\text{if $action$ is not appropriate}

\end{cases}}
\end{equation}

\begin{algorithm}[!ht]
\caption{Forecasting Environment} \label{alg1:cap}
\begin{algorithmic}[1]
\Require{}{time series data $\mathcal{D}=\{s_{t-n},..,s_{t-2},s_{t-1},s_{t},\}$}; {a set of labels $\mathcal{K}=\{1,2,\dots,K\}$}\vfill
\Ensure{} Predicted  time series data  of $\mathcal{K}$, a set of labels, in the form of states $\{s_{t+1},s_{t+2},s_{t+3},s_{t+4}\}$.\vfill
\State Define $forecast\_model \leftarrow Bi\_LSTM Model$
\State $ Train(forecast\_model)\leftarrow forecast\_model(D) $
\State $ \{s_{t+1},s_{t+2},s_{t+3},s_{t+4}\} \leftarrow forecast\_model(predict)$
\State $Initialization:\ observation\_space={s_{t}^{i} \epsilon S}, action\_space={a_{t}\epsilon A}, reward R$
\State $Set\ monitor\_length= N$
\If{$action is appropriate$}
\State $R \leftarrow +reward$
\Else
\State $R \leftarrow -reward$
\EndIf
\State $monitor\_length \leftarrow N-1$
\State $s_{t+1} \leftarrow s_{t}(monitor\_length)$
\If {$N=0$}
\State $done=True$
\Else
\State $done=False$
\EndIf
\State $visualize(a_{t}, R, vital\ signs)$
\State $initial\_state \leftarrow s_{t}[0]$ \Comment{reset environment}
\end{algorithmic}
\end{algorithm}

The algorithm~\ref{alg1:cap} presents the forecasting environment where observation space, action space, and reward policy have been configured based on the predicted states. Lines 1-3 in the algorithm define the deep learning model to train and predict the time series forecasting data. Lines 4-5 initialize the class and set boundaries for the observation space, action space, rewards, and monitoring length. Lines 6-10 explain the reward policy for the actions of the learning agent and how the agents get rewarded either positively or negatively. Lines 11-19 present monitoring length and visualization of agent performance and reset the environment if needed.

\subsection{Learning Agent}
In the proposed PDRL framework, the game learning agent DQN algorithm was used. The algorithm was introduced by Google's DeepMind for playing Atari game to play games by just observing the screen without any training or prior knowledge about those games. In this algorithm, the Q-Learning functions' approximation will be computed using neural networks, and the learning agent gets rewarded based on the neural networks' prediction of the right action for the current state. The reward policy has been discussed in detail in Section~\ref{Environment}. 

\subsubsection{Q-Function Approximation}
The neural networks model used in this study to approximate the rewards has three layers input layer, a hidden layer, and an output layer. The input layer has a node for each vital sign of a state, the output layer has a node for each action in the action space. The model is configured with parameters such as the relu activation function, mean square error as loss function, and Adam optimizer. The model gets trained with the state and its corresponding reward. Upon training, the model is able to predict reward for the current state.

The learning agent performs an action $a_{t} \epsilon A$ for a transition from state $s_{t}$ to $s_{t}^{'}$ and achieves a reward $R$ for the action. In this transition process, the maximum of the Q-function in Equation~\ref{modified_action_value} is calculated, and the discount of the calculated value uses a discount factor $\gamma$ to suppress future rewards and focus on immediate rewards. The discounted future reward is added to the current reward to get the target value. The difference between the current prediction from the neural networks and the calculated target value provides a loss function. The loss function is a deviation of the predicted value from the target value and it can be estimated from Equation~\ref{loss function}. The square of the loss function allows for the punishment of the agent for a large loss value. 

\begin{equation}\label{loss function}
   \displaystyle{ loss = (\underbrace{R+\gamma \cdot max(Q^{\pi^{*}}(s,a))}_{target\_value}- \underbrace{Q^{\pi}(s,a)}_{predicted\_value})^{2}}
\end{equation}

\subsubsection{Memorize and Replay}
A simple neural network has the challenge of limited memory and forgetting previous observations once new observations overwrite them. To retrain the model, previous observations can be stored in an array as an experience $e$ that acts as a memory and appends the current state, action, reward, and next state to the memory at time $t$ as $e_{t}=(s_{t},a_{t},r_{t},s_{t+1})$. A sample of previous observations from the memory is randomly selected to train the neural networks using the replay method. In this study, a batch size of 32 previous observations was to retrain the neural network model.

\subsubsection{Exploration and Exploitation}
Exploration and exploitation are two contradictory concepts in RL where exploration is the selection of actions randomly that have never been performed and exploring more possibilities. Exploitation is to select known actions from existing knowledge and previous experiences to maximize the rewards. To balance exploration and exploitation, there are different strategies such as greedy algorithm, epsilon-greedy algorithm, optimistic initialization, and decaying epsilon-greedy algorithm. This study controls the exploration rate by multiplying decay by the exploration rate. This reduces the number of explorations in the execution as the agents learn the patterns and maximize their rewards to get high scores. While the neural networks model is getting retrained with previous experiences in the replay, the decay gets multiplied by the exploration rate based on how well the agent can predict the right actions. All these parameters are defined as hyper-parameters to DQN learning agents. 

\begin{algorithm}[!ht]
\caption{Learning Agent}\label{alg2:cap}
\begin{algorithmic}[1]
\State Initialize $\gamma, \epsilon, \epsilon_{decay}, \epsilon_{min}, memory=\emptyset, batch\_size $ 
\State Define $model \leftarrow Neural Network Model$ 
\State $memory \leftarrow append(s_{t}, a_{t}, R, s_{t+1})$ 
\If {$np.random.rand < \epsilon$} \Comment{Exploration}
\State $action\_value \gets random(a_{t})$
\Else \Comment{Exploitation}
\State $action\_value \gets model.predict(s_{t})$
\EndIf
\State $minibatch \gets random(memory,batch\_size)$
\For {$s_{t}, a_{t}, R, s_{t+1}, done$ in $minibatch$}
\State $target \gets R$
\If {not done}
\State $target \gets R + \gamma(max(model.predict(s_{t+1})))$
\EndIf
\State $target\_f \gets model.predict(s_{t})$
\State $target\_f[a_{t}] \gets target$
\State $model.fit(s_{t}, target\_f)$
\EndFor
\If {$\epsilon$ $\geq$ $\epsilon_{min}$}
\State {$\epsilon$ *= $\epsilon_{decay}$}
\EndIf
\end{algorithmic}
\end{algorithm}

The methods to perform function approximation, memorize, replay, exploration, and exploitation are enclosed in a Learning Agent algorithm~~\ref{alg2:cap}. Line 1-2 initializes all the hyper-parameters required for the agent and a deep learning model for Q-function approximation. Line 3 explains the memorize and replay part to store neural-network experience where state, action, reward, and next\_state will be stored to retrain the model using the replay method. Lines 4-8 in the algorithm are responsible to predict an action either exploration or exploitation methods. Lines 9-21 explain a batch of previous experiences from memory will be retrieved to process and retrain the neural networks model based on the hyper-parameters defined earlier.

\begin{algorithm}[ht]
\caption{Proposed Multiple Agents Monitoring Framework Implementation}\label{alg3:cap}
\begin{algorithmic}[1]
\Require{\textbf{Input:}
\Statex $\mathcal{C}={1,2,\dots,C}$: set of subjects
\Statex $\mathcal{V}={1,2,\dots,V}$: set of vital signs
\Statex $\mathcal{M}={1,2,\dots,M}$: number of episodes
}
\Ensure{\textbf{Output:} Rewards achieved by Agents in each episode.}
\State $env \leftarrow ForecastingEnvironment()$ \Comment{Algorithm~\ref{alg1:cap}}
\State $agent \leftarrow LearningAgent()$ \Comment{Algorithm~\ref{alg2:cap}}
\For {episode $m \in \mathcal{M}$}
\State $state \leftarrow env.reset()$
\State $score=0$
\For {time in range(timesteps)}
\State $a_{t} \leftarrow agent.action(s_{t})$
\State $s_{t+1}, R, done \leftarrow env.step(a_{t})$
\State $agent.memorize(s_{t}, a_{t}, R, s_{t+1})$
\State $s_{t} \leftarrow s_{t+1}$
\If {done}
\State $print(m, score)$
\State $break;$
\EndIf
\EndFor
\State $agent.replay(batch\_size)$
\EndFor
\end{algorithmic}
\end{algorithm}

The Algorithm ~\ref{alg3:cap} is an extension to the previous two Algorithms~\ref{alg1:cap}~\ref{alg2:cap} and implements the proposed generic PDRL monitoring framework. The inputs of the algorithm are a set of subjects and their vital signs along with the number of episodes the agents have to be trained. The algorithm~\ref{alg3:cap} outputs the learning agents score which is the cumulative sum of rewards achieved in each episode. Lines 1-2 create objects of ForecastingEnvironment and LearningAgent. Lines 3-17 are nested for loops with conditional statements to check if the episode is completed or not. The outer loop is to iterate each episode while resetting the environment to initial values and score to zero. The inner loop is to iterate timesteps which denote the time of the current state and call the methods defined in Algorithm~\ref{alg1:cap}~\ref{alg2:cap} to predict action for the current state, to reward the agent for predicted action, to retrieve next\_state, and to memorize the previous experiences. Finally, the replay method will be called to retain the neural network model with the stored previous experiences. 

\section{Experiment}\label{experiment}
The multi-agent PDRL monitoring framework proposed in this study has been evaluated with experiments on different datasets related to healthcare, traffic, and weather forecasting. In healthcare, vital signs such as heart rate, respiration, and temperature retrieved from a patient are processed into time series data. The forecasting environment is responsible to learn the time series data and forecast future vital signs in the next 15 minutes, 30 minutes, 45 minutes, and 60 minutes. The predicted data is passed to multiple DRL agents with one vital sign for each agent. The agents are responsible to monitor the vital signs in each iteration and take appropriate actions. For this task, each agent gets rewarded as per the reward policy discussed in the previous section~\ref{methods}. As they aim to increase their accumulated rewards, all the agents learn each vital sign pattern of patients and collectively monitor the patient's health status. The isolated DRL agents monitor their vital signs independently and update the corresponding  medical emergency team(MET) at the right time.
\begin{table}[]
\centering
\caption{Proposed Multi-Agent PDRL framework performance is compared with other baseline models}
\label{tab:forecastresults}
% \resizebox{0.9\columnwidth}{!}{%
\begin{tabular}{@{}clrrr@{}}
\toprule
\multicolumn{1}{l}{} &
   &
  \multicolumn{1}{l}{\textbf{MAE}} &
  \multicolumn{1}{l}{\textbf{MAPE}} &
  \multicolumn{1}{l}{\textbf{RMSE}} \\ \midrule
\multicolumn{1}{c}{\multirow{4}{*}{\textbf{ELMA~\cite{li2022elma}}}} &
  \multicolumn{1}{l}{\textbf{15 Min}} &
  \multicolumn{1}{r}{6.2} &
  \multicolumn{1}{r}{13.91} &
  \multicolumn{1}{r}{8.75} \\ \cmidrule(l){2-5} 
\multicolumn{1}{c}{} &
  \multicolumn{1}{l}{\textbf{30 Min}} &
  \multicolumn{1}{r}{6.2} &
  \multicolumn{1}{r}{13.91} &
  \multicolumn{1}{r}{8.75} \\ \cmidrule(l){2-5} 
\multicolumn{1}{c}{} &
  \multicolumn{1}{l}{\textbf{45 Min}} &
  \multicolumn{1}{r}{6.2} &
  \multicolumn{1}{r}{13.91} &
  \multicolumn{1}{r}{8.75} \\ \cmidrule(l){2-5} 
\multicolumn{1}{c}{} &
  \multicolumn{1}{l}{\textbf{60 Min}} &
  \multicolumn{1}{r}{6.13} &
  \multicolumn{1}{r}{13.91} &
  \multicolumn{1}{r}{8.67} \\ \midrule
\multicolumn{1}{c}{\multirow{4}{*}{\textbf{GRU~\cite{ma2021multi}}}} &
  \multicolumn{1}{l}{\textbf{15 Min}} &
  \multicolumn{1}{r}{0.95} &
  \multicolumn{1}{r}{5.47} &
  \multicolumn{1}{r}{1.25} \\ \cmidrule(l){2-5} 
\multicolumn{1}{c}{} &
  \multicolumn{1}{l}{\textbf{30 Min}} &
  \multicolumn{1}{r}{0.95} &
  \multicolumn{1}{r}{5.48} &
  \multicolumn{1}{r}{1.25} \\ \cmidrule(l){2-5} 
\multicolumn{1}{c}{} &
  \multicolumn{1}{l}{\textbf{45 Min}} &
  \multicolumn{1}{r}{0.97} &
  \multicolumn{1}{r}{5.51} &
  \multicolumn{1}{r}{1.27} \\ \cmidrule(l){2-5} 
\multicolumn{1}{c}{} &
  \multicolumn{1}{l}{\textbf{60 Min}} &
  \multicolumn{1}{r}{0.98} &
  \multicolumn{1}{r}{5.5} &
  \multicolumn{1}{r}{1.28} \\ \midrule
\multicolumn{1}{c}{\multirow{4}{*}{\textbf{\begin{tabular}[c]{@{}c@{}}GNN-Based \\ Multi-Agent~\cite{jiang2022internet}\end{tabular}}}} &
  \multicolumn{1}{l}{\textbf{15 Min}} &
  \multicolumn{1}{r}{3.64} &
  \multicolumn{1}{r}{8} &
  \multicolumn{1}{r}{2.46} \\ \cmidrule(l){2-5} 
\multicolumn{1}{c}{} &
  \multicolumn{1}{l}{\textbf{30 Min}} &
  \multicolumn{1}{r}{3.99} &
  \multicolumn{1}{r}{3.47} &
  \multicolumn{1}{r}{2.58} \\ \cmidrule(l){2-5} 
\multicolumn{1}{c}{} &
  \multicolumn{1}{l}{\textbf{45 Min}} &
  \multicolumn{1}{r}{4.33} &
  \multicolumn{1}{r}{4.53} &
  \multicolumn{1}{r}{2.69} \\ \cmidrule(l){2-5} 
\multicolumn{1}{c}{} &
  \multicolumn{1}{l}{\textbf{60 Min}} &
  \multicolumn{1}{r}{5.73} &
  \multicolumn{1}{r}{5.27} &
  \multicolumn{1}{r}{3.09} \\ \midrule
\multicolumn{1}{c}{\multirow{4}{*}{\textbf{\begin{tabular}[c]{@{}c@{}}Multi-Agent\\ PDRL Fraemwork (Ours)\end{tabular}}}} &
  \multicolumn{1}{l}{\textbf{15 Min}} &
  \multicolumn{1}{r}{\textbf{0.44}} &
  \multicolumn{1}{r}{\textbf{2.6}} &
  \multicolumn{1}{r}{\textbf{0.85}} \\ \cmidrule(l){2-5} 
\multicolumn{1}{c}{} &
  \multicolumn{1}{l}{\textbf{30 Min}} &
  \multicolumn{1}{r}{\textbf{0.65}} &
  \multicolumn{1}{r}{\textbf{2.67}} &
  \multicolumn{1}{r}{\textbf{1.05}} \\ \cmidrule(l){2-5} 
\multicolumn{1}{c}{} &
  \multicolumn{1}{l}{\textbf{45 Min}} &
  \multicolumn{1}{r}{\textbf{0.52}} &
  \multicolumn{1}{r}{\textbf{3.17}} &
  \multicolumn{1}{r}{\textbf{0.93}} \\ \cmidrule(l){2-5} 
\multicolumn{1}{c}{} &
  \textbf{60 Min} &
  \textbf{0.53} &
  \textbf{5.39} &
  \textbf{0.95} \\ \bottomrule
\end{tabular}
\end{table}

\subsection{Dataset}

\begin{itemize}
    \item PPG-DaLiA~\cite{reiss2019deep}: The dataset contains physiological and motion data of 15 subjects, recorded from both a wrist-worn device and a chest-worn device, while the subjects were performing a wide range of activities under close to real-life conditions. 
    % \item Student Performance~\cite{cortez2008using}: The dataset under consideration is related to student achievement in secondary education of two Portuguese schools. The data contains information about student grades, demographic, social, and school-related features. The data was collected by using school reports and questionnaires. 
    \item Traffic Dataset~\cite{zhao2019spatial}: The dataset includes 47 features such as a historical sequence of traffic volume for the last 10 sample points, day of the week, hour of the day, road direction, number of lanes, and name of the road. 
    \item Meteorological Data~\cite{cortez2007data}: This dataset aims to predict the area affected by forest fires using meteorological data such as temperature, relative humidity, wind, and rain. The area affected by the fires is transformed using an ln(x+1) function before applying various Data Mining methods. 
\end{itemize}

\subsection{Baseline Models}    
\subsubsection{RL Algorithms}
% Considering the decision-making tasks, the proposed PDRL framework agents' performance is compared with the baseline models.
\begin{itemize}
    \item Existing RL baseline models by Li et al.~\cite{li2022electronic} are deployed to optimize sequential treatment strategies based on EHRs for chronic diseases with DQN. The multi-agent framework results are compared with Q Learning and Double DQN.
    \item Additionally, Guo et al.~\cite{guo2021deep} used RL and a dynamic weight assignment network architecture with TD3 (combination of DDPG, Actor-Critic, and DQN) to recognize human activity.
    \item Yom et al.~\cite{yom2017encouraging} used A2C and PPO algorithms to act as virtual coaches in decision-making and send personalized messages.
\end{itemize}

\subsubsection{Predictive RL Frameworks}
% The datasets used to evaluate the proposed framework are trained to supervised learning algorithms as baseline models to predict heart rate. The baseline models are a combination of both machine learning and deep learning models. All the baseline models in both works are evaluated using Root Mean Square Error (RMSE).

\begin{itemize}
    \item Li et al.~\cite{li2022elma} proposed simultaneous energy-based learning for multi-agent activity forecasting using graph neural networks for activity forecasting based on spatio-temporal data.
    \item Ma et al.~\cite{ma2021multi} proposed a multi-agent driving behavior prediction across different scenarios based on the agent's self-supervised domain knowledge.
    \item Jiang et al.~\cite{jiang2022internet} proposed a study on internet traffic prediction with distributed multi-agent learning using LSTM and gated recurrent unit (GRU). GRU-based distributed multi-agent learning scheme achieved the best performance compared to LSTM.

\end{itemize}

\subsection{Evaluation Metrics}
\textbf{Mean Absolute Error (MAE)} is a commonly used regression metric that measures the average magnitude of errors between the predicted and actual values for a set of data. It is calculated as the average of absolute differences between the predicted and actual values and is expressed as a single value. \textbf{Root Mean Squared Error (RMSE)} is another commonly used regression metric that measures the average magnitude of the differences between the predicted and actual values. RMSE is calculated as the square root of the mean of the squared differences between the predicted and actual values. \textbf{Mean Absolute Percentage Error(MAPE)} is a regression metric that measures the average absolute percentage error between the predicted and actual values. It is calculated as the average of the absolute differences between the predicted and actual values, expressed as a percentage of the actual values. \textbf{Cumulative Rewards} is a performance metric used in reinforcement learning to measure the total rewards obtained by an agent over a specified period of time or number of actions. It is calculated as the sum of all rewards received by the agent over the given period of time or number of actions. 
    % The cumulative reward can be used to evaluate the effectiveness of a reinforcement learning algorithm or to compare different algorithms.
All the experiments were conducted using Python version 3.7.6 and the TensorFlow, Keras, Open AI Gym, and stable\_baselines3 packages.

\section{Experiment Results and Analysis}\label{results}
In this section, we conduct an analysis and comparison of the performance of the deep learning model within the forecasting environment of the proposed framework against baseline models. Additionally, we evaluate and compare the performance of the monitoring RL agents with baseline models. The proposed framework utilizes deep learning for forecasting the states of the RL environment, while the RL agent monitors the forecasted states. This makes RL a suitable approach for monitoring applications, such as health monitoring, weather monitoring, traffic monitoring, and more. Moreover, it enables the automation of monitoring tasks, reducing the reliance on manual intervention and increasing the efficiency of monitoring processes. RL agents can continuously monitor the environment, make real-time decisions, and take appropriate actions, allowing human operators to focus on other critical tasks.

\subsection{Forecasting Environment Results}
Traditional machine learning and deep learning algorithms are capable of predicting heart rate in a supervised learning approach. The baseline models with predicting capability are adopted in the PDRL framework to replace the proposed DQN algorithm. They are trained with a subject from PPG-DaLiA to forecast heart rate based on physiological features. Tab.~\ref{tab:forecastresults} presents the results of various frameworks for time series forecasting in the RL environment. The frameworks being compared are ELMA, GRU, GNN-Based Multi-Agent, and the proposed generic Multi-Agent PDRL Framework. The performance of each framework is evaluated using the three metrics: MAE, MAPE, and RMSE. The results show that the proposed multi-agent PDRL framework performs the best among all the models across all time intervals (15 min, 30 min, 45 min, and 60 min). This can be seen by the lowest values of MAE, MAPE, and RMSE for this model. The GRU model also performs well across all time intervals, with MAE, MAPE, and RMSE values significantly lower than those of the ELMA and GNN-Based Multi-Agent models. It is also worth noting that the performance of the GNN-Based Multi-Agent model is inconsistent across all time intervals, showing varying results for different time steps. The results suggest that the GRU and the proposed Multi-Agent PDRL Framework models, which are specifically designed for time series data, performed much better than ELMA and GNN-Based Multi-Agent which are not well suited for time series forecasting.

% \usepackage{booktabs}
\begin{table}[]
\centering
\caption{Baseline Models Comparison}
\label{tab:baseline}
\begin{tabular}{@{}lrrr@{}}
\toprule
\textbf{RL Method} & \multicolumn{1}{l}{\textbf{Agent 1}} & \multicolumn{1}{l}{\textbf{Agent 2}} & \multicolumn{1}{l}{\textbf{Agent 3}} \\ \midrule
Q Learning   & 25878 & 17304 & 23688 \\ \midrule
PPO          & 23688 & 20367 & 17688 \\ \midrule
A2C          & 24717 & 13707 & 24369 \\ \midrule
Double DQN   & 25569 & 15360 & 20367 \\ \midrule
DDPG         & 26760 & 20754 & 23967 \\ \midrule
\textbf{Proposed DQN (Ours)} & \textbf{48354} & \textbf{30019} & \textbf{38651} \\ \bottomrule
\end{tabular}
\end{table}
\subsection{DRL Agents Performance}
The multi-agent in the proposed PDRL monitoring framework has been evaluated with vital signs such as heart rate, respiration, and temperature predicted in the forecasting environment. The agents are responsible to monitor the vital signs in each iteration and take appropriate actions. For this task, each agent gets rewarded as per the reward policy discussed in the previous section~\ref{methods}. As they aim to increase their accumulated rewards, all the agents learn each vital sign pattern of patients and collectively monitor the patient's health status. The isolated PDRL agents monitor their vital signs independently and update the corresponding MET at the right time.

 All the baseline models and the proposed algorithm were trained with the same client data for 10 episodes and their cumulative rewards achieved in the 10th episode are shown in Tab.~\ref{tab:baseline}. Q-Learning algorithm which updates its action-value Bellman equation stopped far behind in learning the patterns of the vital signs compared to the proposed model-free DQN algorithm. The other baseline models PPO, Actor-Critic, Double DQN, and DDPG have considerable performance in all three monitoring agents but couldn't overshadow the results of the proposed DQN approach. The multi-model algorithm TD3 was able to score closer to the proposed approach. Overall, the proposed PDRL algorithm has outperformed all other baseline models in all three monitoring agents.

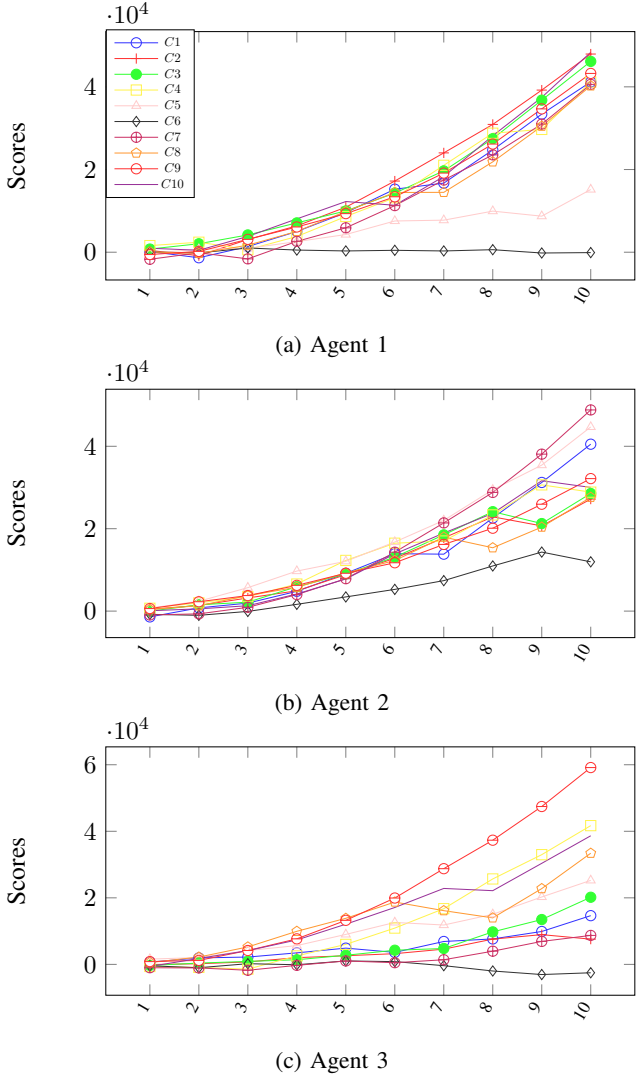
\begin{figure}[ht]
     \centering
     \begin{subfigure}[b]{\columnwidth}
        \centering
        \pgfplotstableread[col sep=comma,]{agent1.csv}\datatable
        \resizebox{\columnwidth}{!}{%
        \begin{tikzpicture}[thick,scale=0.3]
        \begin{axis}[
            width=\columnwidth,
            height=5cm,
            xtick=data,
            xticklabels from table={\datatable}{Episode},
            x tick label style={font=\scriptsize, rotate=60, anchor=east},
            legend style={nodes={scale=0.5, transform shape},at={(0.0, 0.33)},anchor=south west},
            ylabel={Scores}]
            
            \addplot [mark=o, blue!80 ] table [x expr=\coordindex, y={C 1}]{\datatable};
            \addlegendentry{$C 1$}
            
            \addplot [mark=+, red!80 ] table [x expr=\coordindex, y={C 2}]{\datatable};
            \addlegendentry{$C 2$}
            
             \addplot [mark=*, green!80 ] table [x expr=\coordindex, y={C 3}]{\datatable};
            \addlegendentry{$C 3$}
             \addplot [mark=square, yellow!80 ] table [x expr=\coordindex, y={C 4}]{\datatable};
            \addlegendentry{$C 4$}
             \addplot [mark=triangle, pink!80 ] table [x expr=\coordindex, y={C 5}]{\datatable};
            \addlegendentry{$C 5$} 
            
            \addplot [mark=diamond, black!80 ] table [x expr=\coordindex, y={C 6}]{\datatable};
            \addlegendentry{$C 6$}
             \addplot [mark=oplus, purple!80 ] table [x expr=\coordindex, y={C 7}]{\datatable};
            \addlegendentry{$C 7$}
             \addplot [mark=pentagon, orange!80 ] table [x expr=\coordindex, y={C 8}]{\datatable};
            \addlegendentry{$C 8$}
             \addplot [mark=halfcircle, red!80 ] table [x expr=\coordindex, y={C 9}]{\datatable};
            \addlegendentry{$C 9$}
             \addplot [mark=cubes, violet!80 ] table [x expr=\coordindex, y={C 10}]{\datatable};
            \addlegendentry{$C 10$}

        \end{axis}
        \end{tikzpicture}
        }
         \caption{Agent 1}
         \label{fig:agent1_per}
     \end{subfigure}
     \vfill
     \begin{subfigure}[b]{\columnwidth}
         \centering
         \pgfplotstableread[col sep=comma,]{agent2.csv}\datatable
        \resizebox{\columnwidth}{!}{%
        \begin{tikzpicture}[thick,scale=0.3]
        \begin{axis}[
            width=\columnwidth,
            height=5cm,
            xtick=data,
            xticklabels from table={\datatable}{Episode},
            x tick label style={font=\scriptsize, rotate=60, anchor=east},
            legend style={nodes={scale=0.4, transform shape},at={(0.0, 0.43)},anchor=south west}
            ,
            ylabel={Scores}
            ]
            
            \addplot [mark=o, blue!80 ] table [x expr=\coordindex, y={Subject 1}]{\datatable};
            \addlegendentry{$Subject 1$}
             \legend{}
            \addplot [mark=+, red!80 ] table [x expr=\coordindex, y={Subject 2}]{\datatable};
            \addlegendentry{$Subject 2$}
             \legend{}
             \addplot [mark=*, green!80 ] table [x expr=\coordindex, y={Subject 3}]{\datatable};
            \addlegendentry{$Subject 3$}
             \legend{}
             \addplot [mark=square, yellow!80 ] table [x expr=\coordindex, y={Subject 4}]{\datatable};
            \addlegendentry{$Subject 4$}
             \legend{}
             \addplot [mark=triangle, pink!80 ] table [x expr=\coordindex, y={Subject 5}]{\datatable};
            \addlegendentry{$Subject 5$} 
             \legend{}
            \addplot [mark=diamond, black!80 ] table [x expr=\coordindex, y={Subject 6}]{\datatable};
            \addlegendentry{$Subject 6$}
             \legend{}
             \addplot [mark=oplus, purple!80 ] table [x expr=\coordindex, y={Subject 7}]{\datatable};
            \addlegendentry{$Subject 7$}
             \legend{}
             \addplot [mark=pentagon, orange!80 ] table [x expr=\coordindex, y={Subject 8}]{\datatable};
            \addlegendentry{$Subject 8$}
             \legend{}
             \addplot [mark=halfcircle, red!80 ] table [x expr=\coordindex, y={Subject 9}]{\datatable};
            \addlegendentry{$Subject 9$}
             \legend{}
             \addplot [mark=cubes, violet!80 ] table [x expr=\coordindex, y={Subject 10}]{\datatable};
            \addlegendentry{$Subject 10$} 
             \legend{}
        
        \end{axis}
        \end{tikzpicture}
        }
         \caption{Agent 2}
         \label{fig:agent2_per}
     \end{subfigure}
     \vfill
     \begin{subfigure}[b]{\columnwidth}
        
         \centering
         \pgfplotstableread[col sep=comma,]{agent3.csv}\datatable
        \resizebox{\columnwidth}{!}{%
        \begin{tikzpicture}[thick,scale=0.3]
        \begin{axis}[
            width=\columnwidth,
            height=5cm,
            xtick=data,
            xticklabels from table={\datatable}{Episode},
            x tick label style={font=\scriptsize, rotate=60, anchor=east},
            legend style={nodes={scale=0.4, transform shape,font=\footnotesize},at={(0.0, 0.43)},anchor=south west}
            ,
            ylabel={Scores}
            ]
            
            \addplot [mark=o, blue!80 ] table [x expr=\coordindex, y={Subject 1}]{\datatable};
            \addlegendentry{$Subject 1$}
             \legend{}
            \addplot [mark=+, red!80 ] table [x expr=\coordindex, y={Subject 2}]{\datatable};
            \addlegendentry{$Subject 2$}
             \legend{}
             \addplot [mark=*, green!80 ] table [x expr=\coordindex, y={Subject 3}]{\datatable};
            \addlegendentry{$Subject 3$}
             \legend{}
             \addplot [mark=square, yellow!80 ] table [x expr=\coordindex, y={Subject 4}]{\datatable};
            \addlegendentry{$Subject 4$}
             \legend{}
             \addplot [mark=triangle, pink!80 ] table [x expr=\coordindex, y={Subject 5}]{\datatable};
            \addlegendentry{$Subject 5$} 
             \legend{}
            \addplot [mark=diamond, black!80 ] table [x expr=\coordindex, y={Subject 6}]{\datatable};
            \addlegendentry{$Subject 6$}
             \legend{}
             \addplot [mark=oplus, purple!80 ] table [x expr=\coordindex, y={Subject 7}]{\datatable};
            \addlegendentry{$Subject 7$}
             \legend{}
             \addplot [mark=pentagon, orange!80 ] table [x expr=\coordindex, y={Subject 8}]{\datatable};
            \addlegendentry{$Subject 8$}
             \legend{}
             \addplot [mark=halfcircle, red!80 ] table [x expr=\coordindex, y={Subject 9}]{\datatable};
            \addlegendentry{$Subject 9$}
             \legend{}
             \addplot [mark=cubes, violet!80 ] table [x expr=\coordindex, y={Subject 10}]{\datatable};
            \addlegendentry{$Subject 10$} 
             \legend{}

        \end{axis}
        \end{tikzpicture}
        }
         \caption{Agent 3}
         \label{fig:agent3_per}
     \end{subfigure}
        \caption{DQN Agents Performance}
        \label{fig:DQNAgent}
\end{figure}
All three learning agents have been fed with physiological features such as heart rate, respiration, and temperature respectively. Based on the observation space, action space, and reward policy defined for a customized gym environment for human behavior monitoring, the learning agents were run for 10 episodes shown on the x-axis, and the cumulative rewards have been awarded as scores for each episode shown on the y-axis. The performance of each of the learning agents with respect to each input client (Client 1 to Client 10) data can be seen in Fig.~\ref{fig:DQNAgent}. In the results, Agent 1 refers to the heart rate monitoring agent which has a constant increase of scores for each episode for most of the subjects except subjects 5 and 6. The intermittent low scores in Agent 1 performance is due to the exploration rate in DQN learning where the algorithm tries exploring all the actions randomly instead of using neural networks prediction. Similarly, Agent 2 and Agent 3 monitor two other physiological features respiration and temperature respectively. Agent 2 has performed better than the other two agents and achieved consistent scores for all subjects. Out of all agents, Agent 3, temperature monitoring performance is unsatisfactory. This actually drives us back to the data level and found out the data is with a different scale compared to the MEWS~\cite{signscanberra}. Still, Agent 3 achieved high scores in monitoring subjects 9, 8, 4, and 10.

\section{Other Time Series Forecasting Systems}\label{opti}
In the healthcare forecasting and monitoring experiment, vital signs such as heart rate, respiration, and temperature are predicted based on the time series data in the forecasting RL environment and the DRL agents monitored the predicted vital signs to communicate with the appropriate medical emergency team in adverse situations. There are different domains such as traffic and weather where time series forecasting is critical and making sequential decisions are essential. In this study, two such time series forecasting systems are evaluated using the proposed PDRL framework. For these experiments, the monitoring agents and forecasting environment trained in the healthcare experiment are adopted for other time series forecasting applications such as traffic and weather by storing the knowledge gained from the health monitoring application. In the transfer learning process, the traffic dataset~\cite{zhao2019spatial} and meteorological data~\cite{cortez2007data} is used for the evaluation. In the traffic dataset, a DRL agent is deployed for monitoring the traffic forecasting process by customizing the observation space, action space, and rewards in the RL environment. Similarly, a DRL agent is deployed for monitoring the weather forecasting process.

Tab.~\ref{tab:tlforecast} appears to show the results of the transfer learning experiment in which different models (ELMA, GRU, GNN-Based Multi-Agent, the proposed Multi-Agent PDRL Framework) are used to predict traffic and meteorological data. The results are shown in terms of several evaluation metrics: MAE, MAPE, and RMSE. The time intervals (15 min, 30 min, 45 min, 60 min) indicate the time intervals for which the predictions are made. GRU and the proposed Multi-Agent PDRL Framework models are performing the best for both traffic and meteorological data forecasting across all time intervals. The ELMA and GNN-Based Multi-Agent models, on the other hand, do not perform as well as the GRU and the proposed Multi-Agent PDRL Framework models.

% \usepackage{booktabs}
% \usepackage{multirow}
% \usepackage{booktabs}
% \usepackage{multirow}
\begin{table}[]
\caption{Proposed Multi-Agent PDRL framework performance for traffic and weather prediction}
\label{tab:tlforecast}
\scriptsize
\resizebox{\columnwidth}{!}{%
\begin{tabular}{@{}cccccccc@{}}
\toprule
\multicolumn{1}{l}{} &
  \multicolumn{1}{l}{\textbf{}} &
  \multicolumn{3}{c}{\textbf{\begin{tabular}[c]{@{}c@{}}Traffic Data \\ Forecasting\end{tabular}}} &
  \multicolumn{3}{c}{\textbf{\begin{tabular}[c]{@{}c@{}}Meteorological \\ Data Forecasting\end{tabular}}} \\ \midrule
\multicolumn{1}{l}{} &
  \multicolumn{1}{l}{} &
  \multicolumn{1}{c}{\textbf{MAE}} &
  \multicolumn{1}{c}{\textbf{MAPE}} &
  \textbf{RMSE} &
  \multicolumn{1}{c}{\textbf{MAE}} &
  \multicolumn{1}{c}{\textbf{MAPE}} &
  \textbf{RMSE} \\ \midrule
\multirow{4}{*}{\textbf{ELMA~\cite{li2022elma}}} &
  15 Min &
  \multicolumn{1}{c}{6.73} &
  \multicolumn{1}{c}{15.14} &
  9.4 &
  \multicolumn{1}{c}{6.69} &
  \multicolumn{1}{c}{15.02} &
  9.39 \\ \cmidrule(l){2-8} 
 &
  30 Min &
  \multicolumn{1}{c}{6.73} &
  \multicolumn{1}{c}{15.14} &
  9.4 &
  \multicolumn{1}{c}{6.69} &
  \multicolumn{1}{c}{15.02} &
  9.39 \\ \cmidrule(l){2-8} 
 &
  45 Min &
  \multicolumn{1}{c}{6.73} &
  \multicolumn{1}{c}{15.14} &
  9.4 &
  \multicolumn{1}{c}{6.69} &
  \multicolumn{1}{c}{15.02} &
  9.39 \\ \cmidrule(l){2-8} 
 &
  60 Min &
  \multicolumn{1}{c}{6.72} &
  \multicolumn{1}{c}{15.07} &
  9.39 &
  \multicolumn{1}{c}{6.65} &
  \multicolumn{1}{c}{14.99} &
  9.34 \\ \midrule
\multirow{4}{*}{\textbf{GRU~\cite{ma2021multi}}} &
  15 Min &
  \multicolumn{1}{c}{1.04} &
  \multicolumn{1}{c}{6.04} &
  1.36 &
  \multicolumn{1}{c}{1.03} &
  \multicolumn{1}{c}{5.96} &
  1.36 \\ \cmidrule(l){2-8} 
 &
  30 Min &
  \multicolumn{1}{c}{1.04} &
  \multicolumn{1}{c}{6.01} &
  1.36 &
  \multicolumn{1}{c}{1.03} &
  \multicolumn{1}{c}{5.94} &
  1.35 \\ \cmidrule(l){2-8} 
 &
  45 Min &
  \multicolumn{1}{c}{1.04} &
  \multicolumn{1}{c}{5.96} &
  1.36 &
  \multicolumn{1}{c}{1.04} &
  \multicolumn{1}{c}{5.93} &
  1.36 \\ \cmidrule(l){2-8} 
 &
  60 Min &
  \multicolumn{1}{c}{1.04} &
  \multicolumn{1}{c}{6.1} &
  1.36 &
  \multicolumn{1}{c}{1.04} &
  \multicolumn{1}{c}{6.02} &
  1.36 \\ \midrule
\multirow{4}{*}{\textbf{\begin{tabular}[c]{@{}c@{}}GNN-Based \\ Multi-Agent~\cite{jiang2022internet}\end{tabular}}} &
  15 Min &
  \multicolumn{1}{c}{4.64} &
  \multicolumn{1}{c}{6.07} &
  2.88 &
  \multicolumn{1}{c}{4.27} &
  \multicolumn{1}{c}{7.32} &
  2.76 \\ \cmidrule(l){2-8} 
 &
  30 Min &
  \multicolumn{1}{c}{5.79} &
  \multicolumn{1}{c}{5.7} &
  3.21 &
  \multicolumn{1}{c}{5.03} &
  \multicolumn{1}{c}{4.71} &
  2.99 \\ \cmidrule(l){2-8} 
 &
  45 Min &
  \multicolumn{1}{c}{6.57} &
  \multicolumn{1}{c}{4.93} &
  3.43 &
  \multicolumn{1}{c}{5.61} &
  \multicolumn{1}{c}{4.89} &
  3.16 \\ \cmidrule(l){2-8} 
 &
  60 Min &
  \multicolumn{1}{c}{7.01} &
  \multicolumn{1}{c}{6.07} &
  3.54 &
  \multicolumn{1}{c}{6.57} &
  \multicolumn{1}{c}{5.86} &
  3.43 \\ \midrule
\multirow{4}{*}{\textbf{\begin{tabular}[c]{@{}c@{}}Multi-Agent\\ PDRL Fraemwork (Ours)\end{tabular}}} &
  15 Min &
  \multicolumn{1}{c}{0.47} &
  \multicolumn{1}{c}{2.94} &
  0.91 &
  \multicolumn{1}{c}{0.44} &
  \multicolumn{1}{c}{2.83} &
  0.89 \\ \cmidrule(l){2-8} 
 &
  30 Min &
  \multicolumn{1}{c}{0.56} &
  \multicolumn{1}{c}{5.61} &
  1.01 &
  \multicolumn{1}{c}{0.6} &
  \multicolumn{1}{c}{4.21} &
  1.04 \\ \cmidrule(l){2-8} 
 &
  45 Min &
  \multicolumn{1}{c}{0.46} &
  \multicolumn{1}{c}{4.04} &
  0.91 &
  \multicolumn{1}{c}{0.48} &
  \multicolumn{1}{c}{3.69} &
  0.92 \\ \cmidrule(l){2-8} 
 &
  60 Min &
  \multicolumn{1}{c}{0.58} &
  \multicolumn{1}{c}{5.71} &
  1.01 &
  \multicolumn{1}{c}{0.54} &
  \multicolumn{1}{c}{5.71} &
  0.99 \\ \bottomrule
\end{tabular}}
\end{table}

\begin{figure}[ht] 
        \pgfplotstableread[col sep=comma,]{tlresult.csv}\datatable
        \resizebox{\columnwidth}{!}{%
        \begin{tikzpicture}[thick,scale=0.6]
        \begin{axis}[
            width=\columnwidth,
            height=5cm,
            xtick=data,
            xticklabels from table={\datatable}{Episode},
            x tick label style={font=\scriptsize, rotate=60, anchor=east},
            legend style={nodes={scale=0.8, transform shape},at={(0.0, 0.43)},anchor=south west},
            ylabel={Scores}]
            
            \addplot [mark=o, blue!80 ] table [x expr=\coordindex, y={Traffic}]{\datatable};
            \addlegendentry{$Traffic$}
            
            \addplot [mark=+, red!80 ] table [x expr=\coordindex, y={Weather}]{\datatable};
            \addlegendentry{$Weather$}
        \end{axis}
        \end{tikzpicture}
        }
         \caption{PDRL Agent results on traffic and weather forecasting}
         \label{fig:tlresult}
     \end{figure}
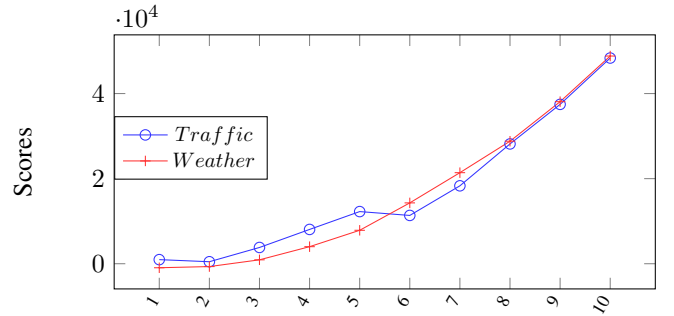

The performance of the proposed PDRL monitoring agents in traffic and weather monitoring applications is presented in Fig.~\ref{fig:tlresult}. It appears that the PDRL agent is able to perform well on both tasks, as the total rewards for both tasks increase with episode number. However, it is also apparent that the agent performs better on the traffic monitoring task than on the weather monitoring task, as the total rewards for the traffic monitoring task are consistently higher than those for the weather monitoring task. Additionally, the gap between the rewards for the two tasks is increasing as the episode number increases, indicating that the agent is becoming increasingly better at the traffic monitoring task.

\section{Discussion}\label{discussion}

The primary objective of the proposed study is to design a multi-agent framework to monitor the predicted future states of a dynamic and complex environment. The proposed framework has been adopted to monitor patients in an uncertain hospital environment where the patient's vital signs fluctuate intermittently and might cause health deterioration with delay in treatment. To overcome this challenge, the sequential decision-making capability of RL algorithms was adopted in this study. Each vital sign in the human body has different threshold levels to determine the health emergency as per MEWS~\cite{signscanberra} and medical emergency teams are predefined for each emergency based on the threshold of the vital sign. In this study, a PDRL agent was deployed for each vital sign and three PDRL agents interacted with the same generic healthcare monitoring environment. The PDRL agents have no prior training or knowledge about patients' vital signs. Based on the reward policy defined in the forecasting environment, the DRL agents learning agents predicted the right action or right MET to communicate the emergency of each vital sign. While designing the environment, setting up the observation space for each DRL agent was critical as it would directly affect the agent learning process and might lead to ambiguity in communicating to the right MET. In this study, PDRL agent 3 was deployed to monitor patients' body temperature, and its performance was unsatisfactory compared to the other two agents. This raises the question of the sanity of input data and observation space configured in the environment. The issue was the units of the temperature thresholds in the MEWS table and the input body temperature of data from the dataset are different. The proposed multi-agent PDRL framework is a generic framework that can be adapted to different time series forecasting applications and monitored to make sequential decisions. In this study, the experiments on healthcare, traffic, and weather data showed promising results. An added advantage of the proposed framework is multi-agents for multiple monitoring parameters in a dynamic environment. This avoids the sparse rewards challenge and can be easily customized and adapted to different applications.

% Based on the results from the forecasting environment, patients' vital signs can be predicted with minimal error rates. However, the supervised learning approach needs external supervision to coordinate the learning process and communicate the information to the right team for timely intervention. Other than the external supervision, the data modeling stage demands a huge amount of labeled data which requires a massive amount of time and resources. For instance, the 10 subjects' data used to evaluate this study was labeled based on the ground truth information from the ECG signal and identified R-peaks are processed to extract the label heart rate. Further to this, the model-building and fine-tuning tasks also require dedicated resources. All of these processes undertake to assist clinicians in the decision-making process with informed decisions. 

% With the ability to tackle decision-making tasks by deploying learning agents, RL can assist clinicians by communicating patient health status in a real-time manner. The proposed DRL-based monitoring framework requires a one-time design of an environment with observation space, action space, and reward definition that can form the ground for the monitoring process. The learning agent learns the patterns of patients' features and monitors their vital signs with any prior knowledge or information. The RL approach with virtual agents is more efficient in monitoring patients without affecting daily activities in a hospital, in-home, or specialized health facilities.

\section{Conclusion}\label{conclusion}
This study proposes a new paradigm of monitoring forecasted states using multiple DRL agents. A generic PDRL monitoring environment was designed with a reward policy to reward the DRL agents based on their actions in each iteration of monitoring status. The learning agents were compelled to learn the behavior of the data patterns based on the reward policy for all possible actions in the action space for each state in the continuous observation space. Based on the evaluation results, all three DRL agents in the PDRL framework were able to learn the patterns of the vital signs and predict appropriate action to alert corresponding medical emergency teams. Furthermore, the knowledge from health monitoring is stored and performed transfer learning process on traffic and weather monitoring. However, the limitation of this study is the input data scale, or units of states, that the agent is monitoring. This led to the under-performance of DRL agent 3 compared to the other two DRL agents in the health monitoring application. Ensemble methods, such as combining the predictions from multiple DRL agents or combining the PDRL framework with other machine learning approaches, could be explored to further improve the accuracy and robustness of the forecasting and decision-making capabilities of the system.
% The proposed three model-free learning gaming DRL agents have outperformed other RL agents in the baseline works related to decision-making tasks.

% \section*{Conflict of Interest}
% The authors have no conflicts of interest to declare that are relevant to the content of this article.

% \section*{Data Availability}
% This study utilized publicly available datasets to evaluate the proposed framework. The following datasets were used:
% \begin{itemize}
%     \item PPG-DaLiA~\cite{reiss2019deep}
%     \item Traffic Dataset~\cite{zhao2019spatial}
%     \item Meteorological Data~\cite{cortez2007data}
% \end{itemize}

\bibliographystyle{ieeetr}
\bibliography{ieee}

% \bibliography{sn-bibliography}% common bib file

\end{document}